\documentclass[acmtog]{acmart}
\acmSubmissionID{640}

\usepackage{booktabs} 

\usepackage{subcaption}
\usepackage{enumitem}

\usepackage[ruled]{algorithm2e} 

\SetAlFnt{\small}
\SetAlCapFnt{\small}
\SetAlCapNameFnt{\small}
\SetAlCapHSkip{0pt}

\copyrightyear{2026}
\acmYear{2026}
\setcopyright{cc}
\setcctype{by}
\acmConference[SIGGRAPH Conference Papers '26]{Special Interest Group on Computer Graphics and Interactive Techniques Conference Conference Papers}{July 19--23, 2026}{Los Angeles, CA, USA}
\acmBooktitle{Special Interest Group on Computer Graphics and Interactive Techniques Conference Conference Papers (SIGGRAPH Conference Papers '26), July 19--23, 2026, Los Angeles, CA, USA}
\acmDOI{10.1145/3799902.3811110}
\acmISBN{979-8-4007-2554-8/2026/07}

\begin{document}
\title{MoonSplat: Monocular Online Gaussian Splatting with $\text{Sim}(3)$ Global Optimization}

\author{Guo Pu}
\orcid{https://orcid.org/0000-0001-7833-907X}
\affiliation{%
 \institution{Wangxuan Institute of Computer Technology, Peking University}
 \country{China}}
\email{guopu@pku.edu.cn}

\author{Yixuan Han}
\orcid{}
\affiliation{%
 \institution{Wangxuan Institute of Computer Technology, Peking University}
 \country{China}}
\email{2501112153@stu.pku.edu.cn}

\author{Haofeng Li}
\orcid{}
\affiliation{%
 \institution{Beijing Hydrogen Intelligent Tech. Co., Ltd.}
 \country{China}}
\email{Lihaofeng@qyzn.com.cn}

\author{Yao Zhang}
\orcid{}
\affiliation{%
 \institution{Beijing Hydrogen Intelligent Tech. Co., Ltd.}
 \country{China}}
\email{zhangyao@qyzn.com.cn}

\author{Hui Zhou}
\orcid{}
\affiliation{%
 \institution{Beijing Hydrogen Intelligent Tech. Co., Ltd.}
 \country{China}}
\email{zhouhui@qyzn.com.cn}

\author{Zhouhui Lian}
\authornote{Corresponding author}
\orcid{https://orcid.org/0000-0002-2683-7170}
\affiliation{%
 \institution{Wangxuan Institute of Computer Technology, Peking University}
 \country{China}}
\email{lianzhouhui@pku.edu.cn}

\renewcommand{\shortauthors}{Pu, et al.}

\begin{abstract}
Online 3D reconstruction from monocular image sequences is a challenging and ongoing research topic. 3D Gaussian Splatting (3DGS), leveraging its high-quality real-time rendering capability, empowers online 3D reconstruction to represent dense scenes with enhanced expressiveness, and thus holds great promise for a wide range of applications such as robotics and AR/VR. However, existing online 3DGS methods still suffer from some key challenges: fragile camera pose estimation due to the lack of global optimization, and low optimization efficiency in large-scale or long-sequence scenarios. To address these issues, we propose a robust and efficient online voxelized 3DGS reconstruction framework integrated with global $\text{Sim}(3)$ optimization, which enables reliable camera tracking and efficient global loop closure for both camera poses and voxelized 3DGS.
To accelerate the convergence of the voxelized 3DGS, we further introduce a color residual learning strategy, which not only boosts optimization speed but also enhances rendering quality. 
Extensive experiments on diverse indoor and outdoor datasets demonstrate that our method achieves state-of-the-art performance in both camera pose estimation accuracy and rendering quality, while retaining real-time efficiency.
Additionally, we develop and deploy a real-world UAV-based active reconstruction system grounded on our proposed method, validating its robustness and generalizability for practical online 3D reconstruction tasks. 
Our code and data are available at \url{https://github.com/TrickyGo/MoonSplat}.
\end{abstract}

%
%


%
%


\begin{teaserfigure}
  \includegraphics[width=\textwidth]{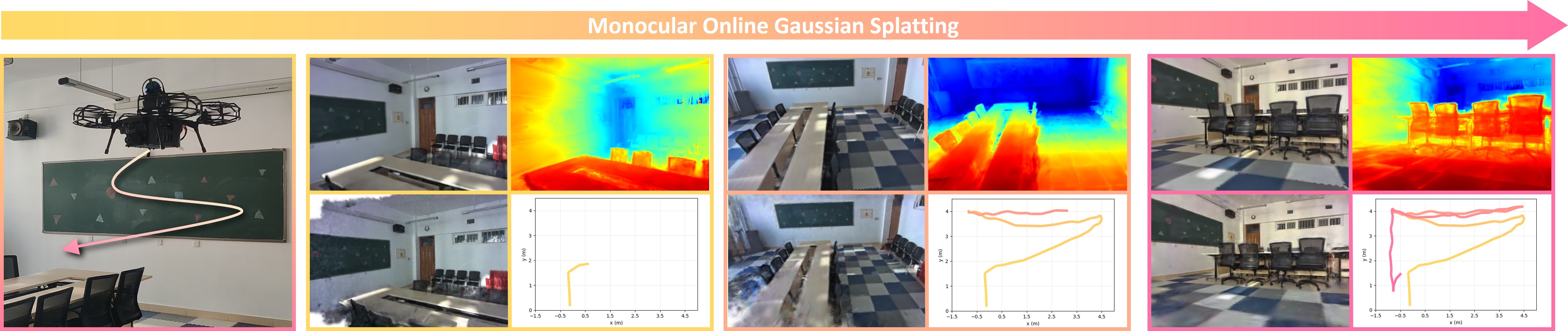}
    \caption{Our method enables real-time reconstruction of online voxelized Gaussian maps with only image stream input. Here we illustrates an example of real-world physical monocular UAV active reconstruction using our method. The left subfigure shows our UAV flying in the scene planning its own trajectory while achieving real-time monocular scene reconstruction. The other three grids present intermediate results during the reconstruction process where in each grid the two upper images shows the RGBD renderings from the current captured viewpoint. The lower left image in each grid is the third-person view rendering, i.e., rendered by a virtual camera placed 0.5 meters behind the current camera pose, which provides a macroscopic observation of the scene. The lower right image in each grid is the horizontal plane component map of the estimated camera poses, where the color gradient from light to dark represents the temporal progression. Since the rendered RGB is highly consistent with the captured RGB, the display of the captured RGB is omitted here.}
  \label{fig:teaser}
\end{teaserfigure}

\maketitle

\section{Introduction}
\label{sec:introduction}

Online 3D reconstruction from monocular image sequences is a long-standing and fundamental challenge in the areas of Computer Graphics and Computer Vision, with extensive applications in robotics, augmented reality/virtual reality (AR/VR), and autonomous driving. 
Traditional online 3D reconstruction methods, particularly simultaneous localization and mapping (SLAM) systems~\cite{davison2007monoslam}, mainly adopt 3D representations such as point clouds, meshes, and signed distance functions (SDFs). 
In recent years, 3D Gaussian Splatting (3DGS)~\cite{kerbl2023} has emerged as a promising representation that offers real-time photo-realistic rendering and differentiable optimization directly from images by modeling scenes as collections of 3D Gaussian primitives. 

Despite its advantages, existing online 3DGS methods~\cite{matsuki2024gaussian, meuleman2025, cheng2025outdoor, deng2025gigaslam} still face critical challenges that hinder their robustness and scalability. 
First, camera pose estimation in existing online 3DGS methods—based on sequential Perspective-n-Point (PnP) pose solving or differential pose optimization via rendering loss—is fragile and prone to failure when camera baselines are insufficient. These methods also lack an effective mechanism to correct cumulative errors in sequential pose estimation and Gaussian maps, such as loop closure correction. While negligible for short sequences, these errors accumulate with increasing sequence length and ultimately lead to tracking failure, resulting in severe reconstruction degradation when loop closures occur. As validated in our experiments, this issue causes non-negligible failure rates on benchmark datasets like ScanNetV2~\cite{dai2017scannet} and Tank-and-Temples~\cite{knapitsch2017tanks}. 
Another challenge is the rapid growth of 3D Gaussian primitives in long sequences, which gives rise to out-of-memory (OOM) issues. To mitigate GPU memory constraints, OTF-NVS~\cite{meuleman2025} stores historical Gaussian subsets to disk as anchors to alleviate OOM. However, this approach prevents these historical Gaussian points from participating in fine-grained global optimization and loop closure correction, further compromising reconstruction consistency. 
GigaSLAM~\cite{deng2025gigaslam} adopts Scaffold-GS~\cite{lu2024scaffold}, a voxelized grid framework where a set of shared multi-layer perceptron (MLP) models predicts Gaussian parameters. Although this reduces memory usage, the joint optimization of grid feature vectors and the shared MLP during online training leads to prohibitively slow optimization, bottle-necking the overall pipeline throughput.

To address these challenges, we propose a robust and efficient online 3DGS reconstruction framework. We leverage pre-trained multi-view 3D priors for robust pose estimation, alleviating reliance on sufficient camera baselines. To continuously correct estimated errors of both camera poses and scale shifts of added Gaussian primitives, we design a $\text{Sim}(3)$ global optimization module to jointly refine camera poses and 3D Gaussian parameters, enabling reliable loop closure and global optimization. 
To avoid memory issues inherent to the original 3DGS~\cite{kerbl2023}, we adopt voxelized 3DGS following Scaffold-GS~\cite{lu2024scaffold} as our scene representation. 
To accelerate the optimization of voxelized 3DGS, we propose a color residual learning strategy that significantly boosts convergence speed while enhancing rendering quality. 
Extensive experiments on various indoor and outdoor scenes demonstrate that our method achieves state-of-the-art performance in camera pose estimation accuracy and rendering quality while maintaining real-time efficiency. 
Additionally, we develop and deploy a real-world UAV-based active reconstruction system grounded on our proposed method, validating its robustness and generalizing ability for practical online 3D reconstruction tasks.
In summary, the main contributions of this paper are fourfold:

\begin{itemize}[topsep=2pt, partopsep=1pt, itemsep=2pt]
    \item We propose an online voxelized 3DGS reconstruction framework integrated with $\text{Sim}(3)$ global optimization, which achieves reliable camera tracking and efficient global loop closure for both camera poses and voxelized 3D Gaussian representations. \looseness=-1
    \item We introduce a color residual learning strategy tailored for voxelized 3DGS, which significantly accelerates optimization convergence. \looseness=-1
    \item Extensive experimental evaluations on diverse indoor and outdoor datasets confirm that our method outperforms state-of-the-art online 3DGS approaches, while retaining real-time performance. \looseness=-1
    \item Based on our method, we develop and deploy a real-world UAV active reconstruction system, validating the robustness, generalizability, and practical utility of our method. \looseness=-1
\end{itemize}

\section{Related Work}
\label{sec:related_work}

\subsection{Visual SLAM}
Visual SLAM aims to infer camera poses and 3D scene geometry from images streams~\cite{cadena2016past}. Traditional SLAM systems are typically formulated as joint optimization problems, with representative sparse monocular SLAM methods such as ORB-SLAM~\cite{murartal2015orb, murartal2017orb2, campos2021orb3} achieving high accuracy in scenes with sufficient features and parallax by optimizing camera pose and sparse 3D landmarks. However, these sparse methods lack dense scene models critical for downstream tasks such as planning and navigation.
To achieve dense reconstruction, researchers have integrated deep learning into SLAM frameworks. Examples include CodeSLAM~\cite{bloesch2018codeslam}, which employs an autoencoder for dense monocular SLAM~\cite{lindenberger2021pixel, tang2018ba}, and DROID-SLAM~\cite{teed2021droid}, which embeds Dense Bundle Adjustment into the optical flow estimation pipeline. 

With the rapid advancement of 3D reconstruction priors (e.g., DUSt3R~\cite{wang2024dust3r}, MAST3R~\cite{leroy2024}, VGGT~\cite{wang2025vggt}, DA3~\cite{lin2025depth}) in Structure-from-Motion (SfM) techniques, state-of-the-art SLAM systems are built upon these priors. For instance, MAST3R-SLAM~\cite{murai2025mast3r} leverages the two-view 3D reconstruction priors MAST3R for tracking, mapping, and re-localization, while VGGT-SLAM~\cite{maggio2025vggt} and VGGT-Long~\cite{deng2025vggt} employs submap alignment for multi-view 3D reconstruction prior VGGT predictions. These state-of-the-art SLAM systems typically involve global camera pose optimization, thus achieving better camera tracking performance than existing online 3DGS methods~\cite{matsuki2024gaussian, meuleman2025, cheng2025outdoor, deng2025gigaslam}.
Despite their advancements, these SLAM systems based on 3D foundation models suffer from a critical limitation: they fail to generate coherent dense 3D representations due to the lack of fine-grained global geometric optimization. Specifically, they cannot fully align point clouds predicted by 3D foundation models, as large-scale bundle adjustment of dense point clouds is indispensable for online SLAMs. While 3DGS brings new opportunities for feasible global geometric optimization in SLAM, 3D reconstruction priors such as MAST3R~\cite{leroy2024} and VGGT~\cite{wang2025vggt} cannot predict robust, consistent camera intrinsics—an essential requirement for rendering methods like NeRF~\cite{barron2021mipnerf} or 3DGS~\cite{kerbl2023}.

Inspired by MAST3R-SLAM~\cite{murai2025mast3r}, we design a $\text{Sim}(3)$ global optimization framework to refine both point clouds and 3D Gaussian parameters, enabling global geometric optimization and high-quality rendering. To estimate accurate camera intrinsics for 3DGS, we develop a camera intrinsics estimation module based on Bundle Adjustment using MAST3R predictions.

\subsection{Online 3DGS Methods}

NeRF-based SLAM methods~\cite{barron2021mipnerf, sucar2021imap, zhu2022nice, zhang2023go, li2026ecslam} achieve photo-realistic reconstruction but remain computationally expensive due to per-ray volumetric rendering, making them unsuitable for real-time online applications. 3DGS~\cite{kerbl2023}, with its superior trade-off between representation richness and real-time efficiency, has become a preferred choice for online 3D reconstruction, spurring extensive integration with SLAM.

Early adaptations include SplaTAM~\cite{keetha2024splatam}, which uses Gaussian primitives and differentiable contour-guided optimization for joint tracking and mapping, and MonoGS~\cite{matsuki2024gaussian}, which introduces geometric verification for dense reconstruction supporting RGB/RGBD inputs but exhibits reduced quality for RGB-only data. Subsequent works such as GS-SLAM~\cite{yan2024gs} propose adaptive expansion for efficient map updates. 

To improve camera pose estimation, Splat-SLAM~\cite{sandstrom2024splat} and Hi-SLAM2~\cite{zhang2025hi} leverages DROID-SLAM, whereas SEGS-SLAM~\cite{wen2025segs} builds upon Orb-SLAM3.
Methods like S3PO-GS~\cite{cheng2025outdoor} and ARTDECO~\cite{li2025artdeco} leverage MAST3R to enhance pose estimation robustness but lack global loop closure for 3DGS, making them prone to failure in long, complex sequences. LoopSplat~\cite{zhu2025loopsplat} establishes a rigid transformation loop closure mechanism for 3DGS but requires RGB-D input.
GigaSLAM~\cite{deng2025gigaslam} attempts global loop closure adjustment via rigid transformations for automotive scenarios but relies on accurate metric depth predictions. However, in many scenes where metric depth predictions are unreliable, this method suffers from tracking failure. To address memory and efficiency issues of the original 3DGS, OTF-NVS~\cite{meuleman2025} uses disk-stored Gaussian anchors, which limits participation in global optimization. Meanwhile, GigaSLAM~\cite{deng2025gigaslam} leverages Scaffold-GS~\cite{lu2024scaffold} to reduce memory usage via voxelized grids but is hampered by slow joint optimization.

To tackle these limitations of existing online 3DGS methods, we propose the first $\text{Sim}(3)$ global optimization for online 3DGS, which directly addresses the inherent scale and pose drift issues of multi-view geometry~\cite{hartley2003multiple}. In addition, our color residual learning strategy accelerates the optimization of voxelized 3DGS, resolving the efficiency bottleneck of voxelized grid-based 3DGS methods for real-time applications.

\section{Method}
\label{sec:method}

\begin{figure*}[!t]
\captionsetup{skip=2pt}
\centering
\includegraphics[width=1\linewidth]{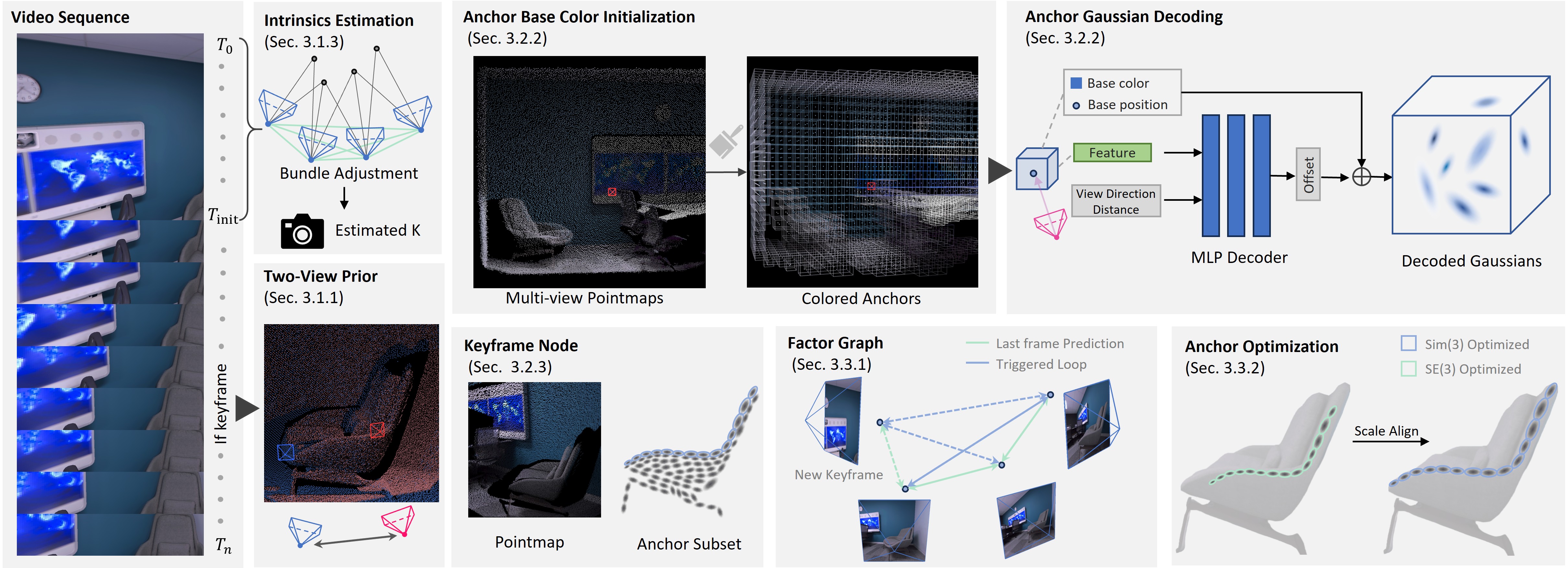}
\caption{Overall pipeline of our proposed online 3DGS framework. Given a monocular image sequence, we first perform camera tracking on each new keyframe. Subsequently, we construct voxelized Gaussians through CRL where gaussian anchors are initialized using the weighted average color of keyframe pointmaps within each voxel, and each voxel is decoded into Gaussian primitives via shared MLPs and \( F_c \) predict color residuals relative to the voxel's base color. Finally, we perform global $\text{Sim}(3)$ optimization based on a factor graph with loop closure to optimize camera poses, scale drifts and synchronously update Gaussian anchor positions, thereby ensuring global scene consistency of the voxelized 3DGS representation.}
\label{fig:pipeline}
\end{figure*}

Our framework takes a monocular image sequence \(\{I_i\}_{i=1}^{N}\) as input. It selects keyframes \(\{I_{k_m}\}_{m=1}^{M}\) where \(m\) is a positive integer satisfying \(1 \leq m \leq N\), and outputs the camera trajectory \(\{R_m, t_m\}_{m=1}^{M}\) including a rotation matrix \(R_m \in \text{SO}(3)\) and a translation vector \(t_m \in \mathbb{R}^3\), along with the 3D scene represented as a voxelized 3D Gaussian Splatting~\cite{lu2024scaffold} \(G_v\). 
For each keyframe \(I_{k_m}\), we perform camera tracking (Sec. 3.1) and mapping (Sec. 3.2), while concurrently performing \(\text{Sim}(3)\) global optimization (Sec. 3.3) to refine both tracking and mapping. The overall pipeline is illustrated in Fig. \ref{fig:pipeline}. \looseness=-1


\subsection{Tracking}
\label{sec:tracking}

\subsubsection{Two-view Prediction}

Our method leverages any state-of-the-art pointmap prediction model to exploit pre-trained multi-view 3D priors for robust pose estimation. In our implementation, we perform two-view matching using MASt3R~\cite{leroy2024}.

MASt3R is a powerful two-view prediction prior which takes a pair of images \(I_i, I_j\), and predicts their corresponding pointmaps \(X_{i,i}, X_{j,i}\) and pointmaps confidence maps \(C_{i,i}, C_{j,i}\), together with matching feature \(D_{i,i}, D_{j,i}\) and matching feature confidences \(Q_{i,i}, Q_{j,i}\), all in the coordinate frame of camera \(i\). Matched points with \(Q > \tau_Q\) are regarded as valid, where \(\tau_Q\) is a threshold for filtering high-confidence matchings.

\subsubsection{Incremental Pose Estimation}

The first frame \(I_1\) is designated as the initial keyframe \(I_{k_1}\), whose pose is set as the identity matrix \(\{R_1, t_1\} = \mathbf{I} \). For each subsequent frame \(I_t\) with \(t > 1\), we first determine whether it qualifies as a keyframe. We first perform MASt3R prediction between \(I_t\) and the latest keyframe \(I_{k_{m}}\). If the ratio of valid matched points falls below a predefined threshold \(\sigma_k\), the current frame \(I_t\) is selected as the new keyframe \(I_{k_{m+1}}\).

Next, we estimate the initial pose of the new keyframe \(I_{k_{m+1}}\) based on \(I_{k_{m}}\) with its pose \(\{R_m, t_m\}\) and existing pointmap \(X_{k_m,k_m}\). We first perform two view prediction with MASt3R and get \(\tilde{X}_{k_{m+1},k_m}\).
Since MASt3R predicts scale independently for each image pair, a rigid \(\text{SE}(3)\) transform consisting of rotation \(R\) and translation \(t\) cannot fully align the \(\tilde{X}_{k_{m+1},k_m}\) and the existing pointmap \(X_{k_m,k_m}\). 
To address the scale misalignment issues, we estimate the relative transform \(T_{k_m,k_{m+1}} \in \text{Sim}(3)\) by minimizing reprojection residuals between \(\tilde{X}_{k_{m+1},k_m}\) and the existing pointmap \(X_{k_m,k_m}\) via a Gauss–Newton solver. The \(\text{Sim}(3)\) transform is parameterized as: 

\[T_{k_m,k_{m+1}} = \begin{bmatrix} s_{k_m,k_{m+1}}R_{k_m,k_{m+1}} & t_{k_m,k_{m+1}} \\ \mathbf{0} & 1 \end{bmatrix},\] 
where \(R_{k_m,k_{m+1}} \in \text{SO}(3)\) denotes the rotation matrix, \(t_{k_m,k_{m+1}} \in \mathbb{R}^3\) denotes the translation vector, and \(s_{k_m,k_{m+1}} \in \mathbb{R}^+\) denotes the scale factor. The \(R_{k_m,k_{m+1}}\) and \(t_{k_m,k_{m+1}}\) components of \(T_{k_m,k_{m+1}}\) are the relative pose between \(I_{k_{m}}\) and \(I_{k_{m+1}}\).
\(X_{k_m+1,k_m+1}\) is initialized as the scale-corrected \(\tilde{X}_{k_{m+1},k_m}\) projected to the coordinate of \(I_{k_{m+1}}\). \looseness=-1

\subsubsection{Camera Intrinsics Estimation}

Pointmaps estimated by MASt3R cannot directly yield stable and accurate camera intrinsics which is a prerequisite essential for 3DGS. To estimate the camera intrinsics, we first perform pairwise matching on the first \(k_{init}\) keyframes using MASt3R to obtain initial relative \(\text{Sim}(3)\) transforms \(T_{p,q}\) (\(1 \le p,q \le k_{init}\)), followed by dense pairwise matching to establish robust keypoint correspondences between all keyframe pairs. 
We then construct a bundle adjustment (BA) problem over this initial keyframe set and matched 3D points from more than two views. We project matched 3D points onto the image plane of each corresponding keyframe using the initial poses \(\{R_{p},t_{p}\}\), and jointly optimize the focal length, 3D points and camera poses via the Levenberg-Marquardt algorithm~\cite{levenberg1944method} to minimize the reprojection error between 2D image keypoints and 3D point projections. 
This optimization yields a unified focal length \(f\) and the camera intrinsics matrix \(K = \begin{bmatrix} f & 0 & c_x \\ 0 & f & c_y \\ 0 & 0 & 1 \end{bmatrix}\), where \((c_x, c_y)\) denotes the principal points. \looseness=-1

To ensure consistency between rendering rays and 3D point positions which is critical for subsequent Gaussian initialization using predicted pointmaps, we correct all subsequent MASt3R-predicted pointmap through depth reprojection. Specifically, we extract the depth component \(z\) from each point in pointmaps, reproject \(z\) onto the camera ray of keyframe using the optimized intrinsic matrix \(K\), and update the 3D coordinates of all points in pointmaps to align with the unified camera intrinsics.
The BA process jointly optimize poses, pointmaps and camera intrinsics, but its computational complexity grows quickly with the number of keyframes, hindering real-time performance. 
Therefore, for subsequent keyframes \(I_{k_{m+1}}\) (\(m \ge k\)), Instead of relying on expensive bundle adjustment, the \(\text{Sim}(3)\) transformations are refined via efficient \(\text{Sim}(3)\) global optimization, as detailed in Sec. 3.3.


\subsection{Mapping}
\label{sec:mapping}

\subsubsection{Voxelized 3DGS}

3DGS~\cite{kerbl2023} enables photo-realistic real-time rendering where a Gaussian primitive \( G \) is parameterized as \( G = (c, s, \alpha, \mu, \Sigma) \), in which \( c \in \mathbb{R}^3 \) denotes the RGB color, \( s \in \mathbb{R}^3 \) is the scale vector, \( \alpha \in \mathbb{R}^+ \) represents opacity, \( \mu \in \mathbb{R}^3 \) is the mean position vector, and \( \Sigma \in \mathbb{R}^{3 \times 3} \) denotes the diagonal covariance matrix. 
For rendering, each 3D Gaussian \( \mathcal{N}(\mu, \Sigma) \) projects onto the image plane as a 2D Gaussian \( \mathcal{N}(\mu_I, \Sigma_I) \), where \( \mu_I = \pi(\xi(\mu)) \), \( \Sigma_I = J \Sigma J^\top \), \( \pi(\cdot) \) is the camera projection function, \( \xi \) is the camera pose, and \( J \) is the Jacobian of the projection transformation and followed distance sort and alpha compositing.
The original 3DGS~\cite{kerbl2023} suffers from prohibitive memory overhead and inefficient optimization due to the rapid growth of Gaussian primitives in large-scale or long-sequence scenes. To mitigate this issue, we adopt the voxelized 3DGS representation following Scaffold-GS~\cite{lu2024scaffold}. Specifically, a voxelized 3DGS \( G_v \) is composed of an anchor set  \( \{a_{u}\}_{u=1}^{U}\) located at a set of voxels and a set of MLPs shared by all anchors for neural Gaussian prediction. Each anchor \( a_{u} \) is associated with a learnable anchor feature \( f_{a_{u}} \in \mathbb{R}^d \) (with \( d \) as the feature dimension). During rendering, \( i \) offset Gaussian primitives are generated per anchor. Shared MLPs predict the offset of each Gaussian relative to the anchor position, along with other Gaussian properties (e.g., color, opacity). For instance, the color MLP \( F_c \) exclusively predicts the RGB color of Gaussian primitives.

\subsubsection{Color Residual Learning}

Voxelized Gaussian is memory-friendly but the joint optimization of anchor features and shared MLPs leads to slow convergence, hindering real-time online performance. We identify that the training of the color MLP is the primary bottleneck: since the anchor feature \( f_{a_u} \) of each anchor is initialized as an empty vector and is jointly trained with Shared MLPs, which means both the anchor features and the color MLP have to learn the color of each anchor from scratch. In the early training stage, invalid color predictions make the optimization of other Gaussian properties ineffective via image loss back-propagation.

\begin{figure*}[!t]
\captionsetup{skip=0pt}
\centering
\includegraphics[width=0.7\linewidth]{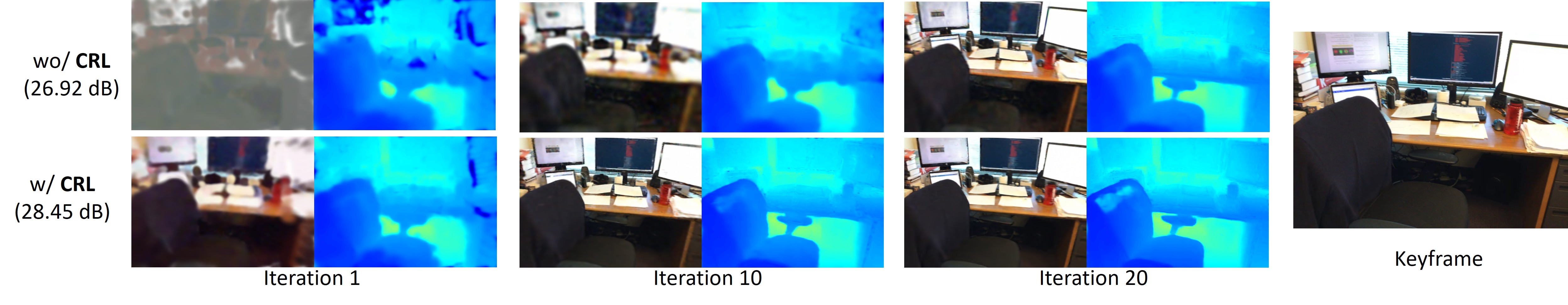}
\caption{\textbf{CRL} accelerates convergence within the same training iterations, leading to better rendering quality. With the anchor base color, CRL produces rendering results close to ground-truth even in the early training stage, avoiding the slow color learning process of the vanilla voxelized 3DGS. }
\label{fig:ablation_color_residual_learning}
\end{figure*}

However, we observe that Gaussian primitives generated from the same anchor exhibit high color similarity, which is also consistent with the color of keyframe pointmaps. Based on this observation, we augment each anchor \( a_u \) with a base color \( c_u \in \mathbb{R}^3 \), computed as the weighted average of RGB colors of keyframe pointmaps falling into the corresponding anchor's voxel. The base color is dynamically updated as new pointmaps are integrated. Leveraging this prior color information, we redesign the color decoding process for Gaussian primitives as Color Residual Learning (CRL), which models the color as a residual relative to the anchor’s base color. In this way, the colored anchor initialized from keyframe pointmaps with colors serves as both geometric and appearance priors. Specifically, we project 3D points with colors from the current keyframe’s pointmap \( X_{k_m,k_m} \) into the global coordinate system using its \( \text{Sim}(3) \) transform \( T_m \). For each voxel, we aggregate the colors and 3D positions of overlapping projected points, assigning the weighted average as the base color \( c_u \) to form the initial colored voxel grid. The weight of each point is determined by its matching confidence \( Q \) to emphasize high-reliability points.
During Gaussian color decoding, the color MLP \( F_c \) predicts a residual color instead of the full RGB color. The input to \( F_c \) is a view-aware fused feature, constructed by concatenating the anchor’s learnable feature \( f_{a_u} \), the unit view direction \( \vec{d}_u \in \mathbb{R}^3 \), defined as the normalized vector from the camera optical center to the anchor \( a_u \), and the voxel-camera distance \( \delta_{u} \in \mathbb{R}^+ \), i.e., the Euclidean distance between the anchor \( a_{u} \) and the camera optical center. The final color of the offset \(i\) Gaussian primitive is given by:

\begin{equation}
c_{u_i} = c_{u} + F_c\left( \left[ f_{a_u}^\top, \vec{d}_u^\top, \delta_u \right]^\top \right), \quad F_c: \mathbb{R}^{d+4} \to \mathbb{R}^3,
\label{eq:color_residual_learning}
\end{equation}
where \( F_c \) consists of 3 fully connected layers with ReLU activation, followed by a tanh layer constraining residuals to \( [-0.5, 0.5]^3 \). Other Gaussian properties (scale \( \mathbf{s} \), opacity \( \alpha \), mean \( \mu \), covariance \( \Sigma \)) retain original prediction pipelines, as color is empirically decoupled from these geometric/transparency attributes.
Fig. \ref{fig:ablation_color_residual_learning} and \ref{fig:ablation_color_residual_learning_2} demonstrates the rendering performance of voxelized Gaussians after the same number of training iterations. With the anchor base color, CRL produces rendering results close to the ground-truth even in the early training stage, avoiding the slow color learning process of the vanilla voxelized 3DGS. In subsequent training, CRL accelerates convergence and improves rendering quality under fixed training iterations. \looseness=-1

\subsubsection{Gaussian Initialization and Training}

For Gaussian initialization in the keyframe \(I_{k_{m}}\), we first acquire the $\text{Sim}(3)$ transform \(T_m\) and pointmap \( X_{k_m,k_m} \) of each keyframe from Sec. 3.1. Following MonoGS~\cite{matsuki2024gaussian}, we select points within \( X_{k_m,k_m} \) that are located in regions with significant image gradient discrepancies between the rendered frame and the keyframe. The selected points form a point cloud for the current keyframe, which is then projected into voxels to generate anchors. Notably, we record the Gaussian anchor subset corresponding to the keyframe \(I_{k_{m}}\) indices \( G_{v_m} \) for subsequent global optimization in Sec. 3.3.

We optimize the Gaussian map using the standard $L_1$ color distance and D-SSIM loss as described in~\cite{kerbl2023}. Additionally, we adopt a depth loss formulated as the $L_1$ loss between the rendered depth and the projected point maps, and an isotropic loss as an $L_1$ regularization term of the scaling parameters for maintaining the geometric regularity of Gaussian primitives.
For each keyframe, we perform \( n_{\text{iter}} \) optimization iterations. With the probability \( p_{\text{train}} \), we use the current keyframe as the training view. Otherwise, a random historical keyframe is sampled for training.

Following GigaSLAM~\cite{deng2025gigaslam}, we use a spatial hash function that maps 3D coordinates to unique hash values, enabling constant-time duplicate detection. Only unique anchors are retained to avoid redundancy. \looseness=-1


\subsection{Global Optimization}
\label{sec:global_opt}

\subsubsection{Factor Graph Construction}

Cumulative pose and scale drift are inevitable in sequential tracking and severely degrade the global structure of 3DGS. This is because 3DGS is composed of point cloud projections derived from historical poses and scales, a factor that impairs the global scene consistency. 
To mitigate this issue, we integrate loop closure detection into the \( \text{Sim}(3) \) global optimization of poses, scales, and 3DGS via a factor graph.
We define the factor graph as a collection of nodes, edges and factors $ G_f = (V_f, E_f, F_f)$. Each node corresponds to a keyframe, storing its RGB image, weighted averaged pointmap, \( \text{Sim}(3) \) transform, and Gaussian anchor subset index. The pointmap is updated via a running weighted average filter whenever new MASt3R-predicted points are generated. Edges connect consecutive nodes or loop-closed nodes and stores point matchings between nodes as well as the corresponding confidence scores. Each edge is associated with a reprojection error factor of matched points, which constrains the relative \( \text{Sim}(3) \) transformation between the connected nodes.

For loop closure detection, following MASt3R-SfM~\cite{duisterhof2025mast3r}, we adapt the Aggregated Selective Match Kernel (ASMK) descriptor~\cite{tolias2013aggregate, tolias2020learning}. Specifically, the MASt3R-encoded features of all keyframes are added to an inverted file index in the retrieval database. When a new keyframe is added, it is first connected to the most recent existing node. We then query the ASMK feature database with its MASt3R-encoded features to retrieve the top-\( k_{\text{loop}} \) most similar historical keyframes. For each retrieved keyframe, if the retrieval score exceeds a threshold \( \omega_r \), we perform MASt3R matching and if the number of valid matches exceeds \( \omega_l \), a bidirectional loop closure edge is added to the factor graph.

\subsubsection{$\text{Sim}(3)$ Optimization}

To perform optimization in the factor graph, we first optimize the \( \text{Sim}(3) \) transforms \( \{T_{m}\}_{m=1}^{M} \) of each nodes by minimizing the aggregated weighted reprojection error between each bidirectional edge:

\begin{equation}
\begin{aligned}
\{\tilde{T}_{m}\}_{m=1}^{M} &= \arg\min_{\{T_{m}\}_{m=1}^{M}} \sum_{(i,j) \in E_f} \sum_{(p_i,p_j) \in P_{i,j}} \frac{1}{C_{ij}} \Bigg[ \left\| \pi(p_j) - \pi\big(T_j^{-1} T_i (p_i)\big) \right\|_2^2 \\
&\quad + \left\| \pi(p_i) - \pi\big(T_i^{-1} T_j (p_j)\big) \right\|_2^2 \Bigg],
\end{aligned}
\label{eq:global_sim3_optimization}
\end{equation}
where \( p \in \mathbb{R}^3 \) denotes a 3D matched point between nodes, \( C \) is the confidence score of a matched point, \( \pi(\cdot) \) is the camera projection function parameterized by \( K \), respectively.
This optimization problem is solved using the Gauss-Newton method with sparse Cholesky decomposition to exploit the graph’s sparsity, ensuring efficiency.

With the optimized \( \{\tilde{T}_{m}\}_{m=1}^{M} \), we further optimize the positions of Gaussian anchor points accordingly. For each node we have the recorded Gaussian anchor subset \(G_{v_m}\) associated with the keyframe \(I_{k_m}\) with their original positions \(\mu_{m}\), original global transform \(T_{m}\), and optimized global transform \(\tilde{T}_{m}\). The optimized position \(\tilde{\mu}_{m}\) is computed as:

\begin{equation}
\tilde{\mu}_{m} = \tilde{T}_{m} \cdot T_{m}^{-1} \cdot \mu_{m}.
\label{eq:gaussian_anchor_update}
\end{equation}

In this way, the historical Gaussian anchors added based on the original poses and scales of past keyframes are updated synchronously with the optimization of historical poses and scales, enabling efficient global consistency optimization. Subsequent to position updates, each Gaussian anchor is reassigned to the nearest voxel to maintain the voxelized 3DGS structure, ensuring consistency between the optimized poses and Gaussian anchors.

Existing Method MAST3R-SLAM~\cite{murai2024mast3rslam} optimizes only \( \text{Sim}(3) \) transformations over filtered weighted-averaged pointmaps, failing to support fine-grained scene alignment.
In contrast, we jointly optimize \( \text{Sim}(3) \) transformations (for camera poses, scale alignment, and 3DGS depth loss) and 3DGS anchor positions for each keyframe. By associating anchors with keyframes and optimizing photometric consistency, our method enables globally consistent, fine-grained alignment and loop closing for 3DGS.

\section{Experiments}

\label{sec:experiments}

\subsection{Comparison with Previous Methods}

\paragraph{Baselines}
We compare with three open-source state-of-the-art methods for online 3D Gaussian Splatting: OTF-NVS~\cite{meuleman2025}, S3PO-GS~\cite{cheng2025outdoor}, and GigaSLAM~\cite{deng2025gigaslam}. Notably, our method and OTF-NVS do not require camera intrinsic parameters as input and both predict intrinsics independently with only image inputs. In contrast, S3PO-GS and GigaSLAM rely on pre-provided camera intrinsics, so we provide these two methods with ground-truth intrinsics. 

\paragraph{Datasets}
We evaluate all methods on 30 diverse indoor and outdoor scenes with complex trajectories and large scales from Tank-and-Temples~\cite{knapitsch2017tanks}, ScanNetV2~\cite{dai2017scannet} and  Waymo~\cite{sun2020scalability}. These scenes feature diverse layouts and textures, posing significant challenges to online reconstruction robustness, and are widely adopted for evaluating 3D reconstruction and novel view synthesis. The number of images per scene ranges from hundreds to thousands. \looseness=-1

\paragraph{Evaluation Metrics}
To measure camera tracking accuracy, we adopt Absolute Trajectory Error (ATE). For the evaluation of rendering quality, we adopt Peak Signal-to-Noise Ratio (PSNR), Structural Similarity Index (SSIM), and Learned Perceptual Image Patch Similarity (LPIPS), to assess visual fidelity and perceptual consistency. 
For each sequence, every 10th frame is used as test frame. All method outputs are upsampled to the original dataset resolution before computing rendering metrics.
As efficiency is important for online methods, we report Frames Per Second (FPS) of each methods on both datasets. \looseness=-1

\begin{figure*}[!t]
\captionsetup{skip=0pt}
\centering
\includegraphics[width=1\linewidth]{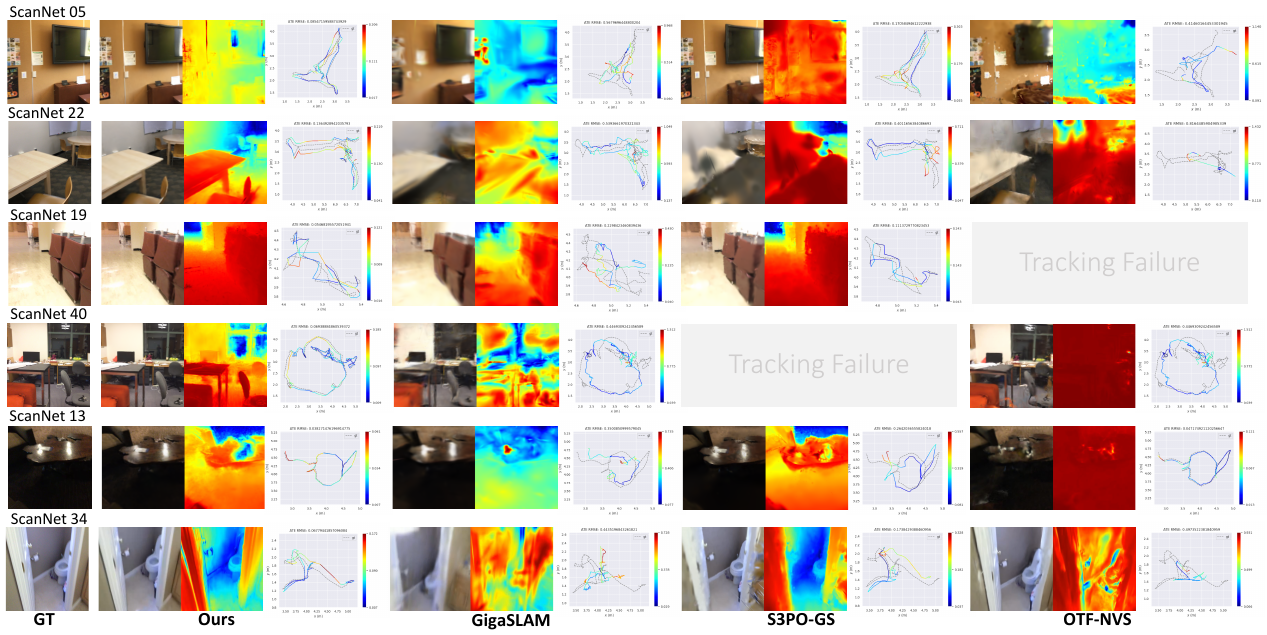}
\caption{Qualitative comparison on the ScanNetV2 dataset.}
\label{fig:Full_comparison_1}
\end{figure*}

\begin{figure*}[!t]
\captionsetup{skip=0pt}
\centering
\includegraphics[width=1\linewidth]{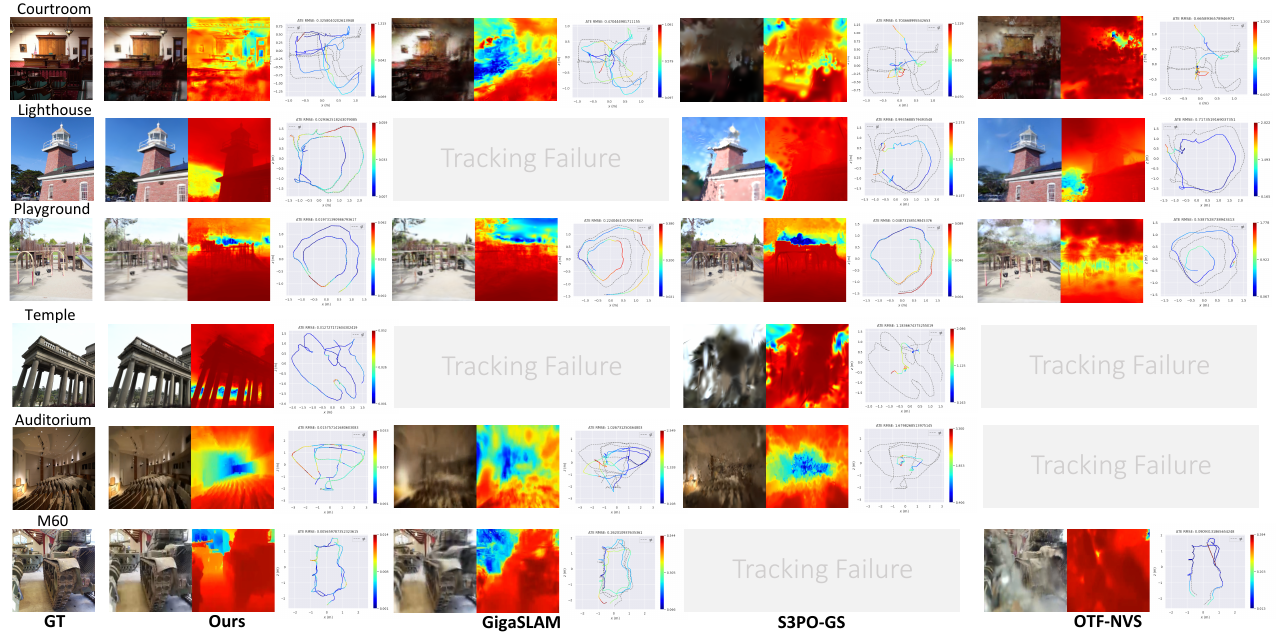}
\caption{Qualitative comparison on the Tank-and-Temples dataset.}
\label{fig:Full_comparison_2}
\end{figure*}

\begin{figure*}[!t]
\captionsetup{skip=0pt}
\centering
\includegraphics[width=1\linewidth]{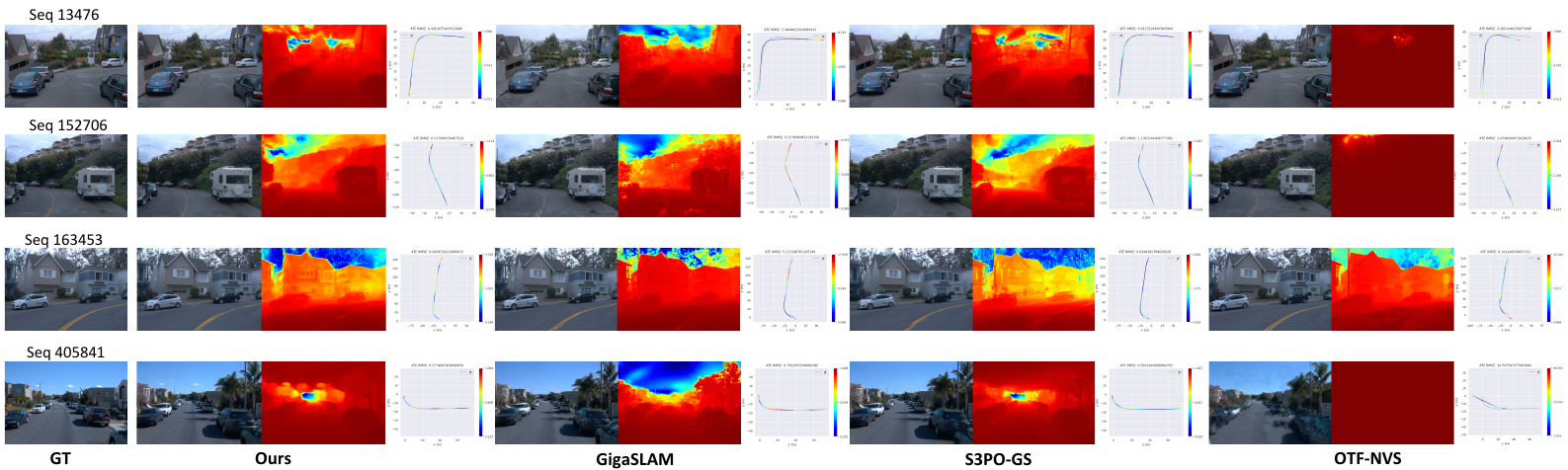}
\caption{Qualitative comparison on the Waymo dataset.}
\label{fig:Full_comparison_3}
\end{figure*}

\paragraph{Qualitative Results}
Fig. \ref{fig:Full_comparison_1}, Fig. \ref{fig:Full_comparison_2} and Fig. \ref{fig:Full_comparison_3} presents qualitative comparisons. Our method captures finer scene details and maintains consistent rendering quality across thousands-frame-long-sequences, while baseline methods suffer from blurring artifacts and drift-induced inconsistencies. This advantage is primarily attributed to the integrated loop closure detection and $\text{Sim}(3)$ global optimization module, which effectively mitigates cumulative pose and scale drift. This ensures stable pose estimation and scene consistency in challenging long-sequence scenarios, while our CRL enhances rendering quality and accelerates convergence of the voxelized Gaussian. \looseness=-1

\paragraph{Quantitative Results}
Quantitative results on Tank-and-Temples, ScanNetV2 and Waymo are shown in Table~\ref{tab:tank_compact},~\ref{tab:scannet_compact} and~\ref{tab:waymo_compact}, respectively. Consistent with qualitative comparisons, our method achieves the highest tracking success rate and outperforms all other methods in comprehensive metrics, including ATE, PSNR, SSIM, and LPIPS. It demonstrates superior robustness and accuracy in both large-scale indoor and outdoor scenes, maintaining reliable performance.

\paragraph{Result Analysis}
On Tanks-and-Temples/ScanNetV2, previous methods frequently suffer from tracking failures due to heavy reliance on epipolar/PnP-RANSAC solvers derived from classic SfM/SLAM pipelines, which are built upon two-view epipolar geometry. Such solvers require a sufficient camera baseline to triangulate keypoints and estimate accurate poses. However, hand-held capture in Tanks-and-Temples and ScanNetV2 often involves rapid camera rotations with inadequate baselines, leading to severe error accumulation and the breakdown of epipolar/PnP-RANSAC solvers. In contrast, the Waymo dataset, captured by vehicle-mounted platforms, exhibits dominant camera translation with minor rotational motion, which ensures stable pose estimation for existing methods.
Our method achieves robust camera tracking across all datasets by incorporating the two-view geometric prior from MAST3R~\cite{murai2025mast3r}, together with the proposed $\text{Sim}(3)$ global optimization which effectively reduces cumulative pose and 3DGS errors, ultimately ensuring global consistency of the reconstruction. \looseness=-1

\begin{figure*}[!t]
\captionsetup{skip=0pt}
\centering
\begin{minipage}[t]{0.45\linewidth}
    \centering
    \includegraphics[width=\linewidth]{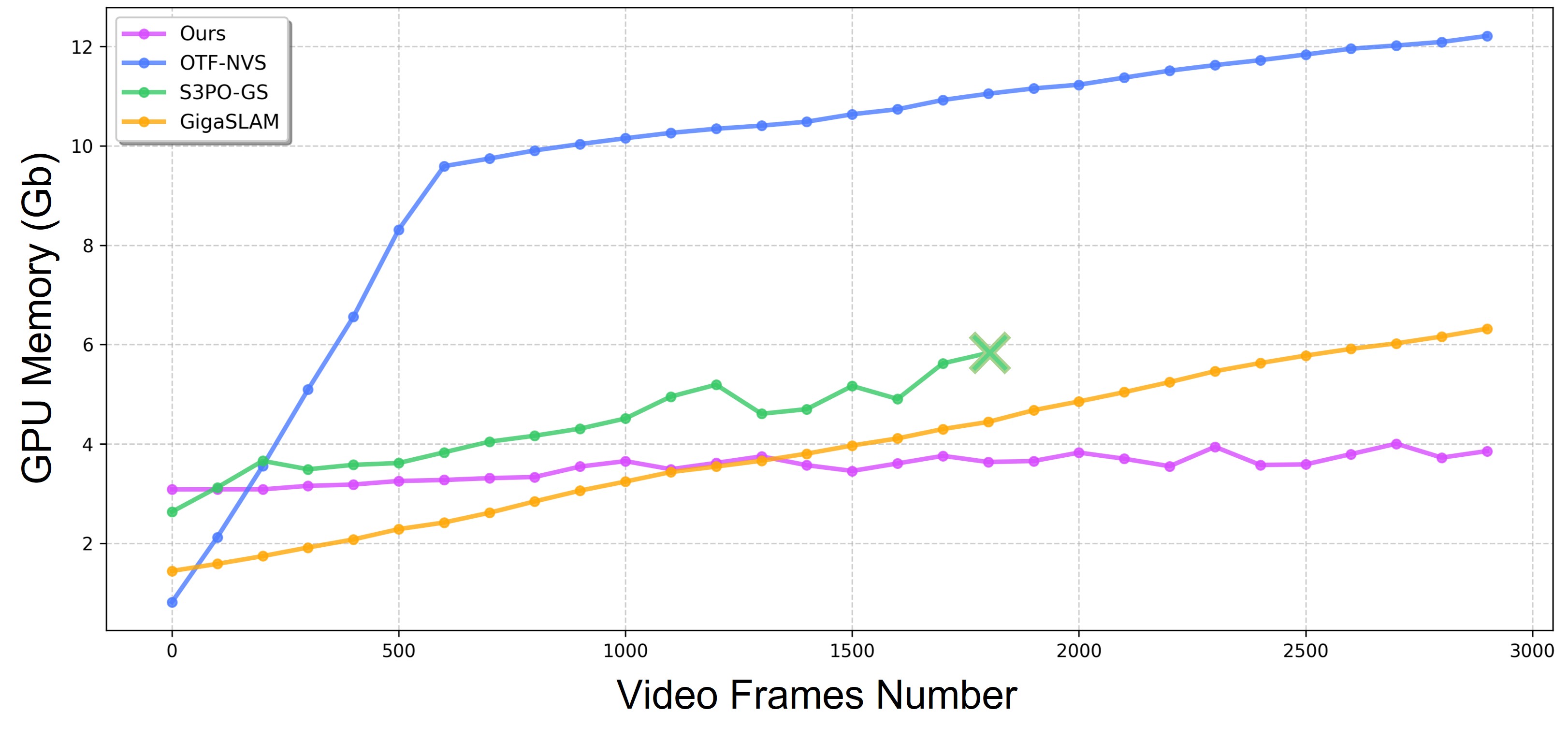}
    \caption{GPU memory footprint comparison. Our method maintains low GPU memory footprint for long image sequences. In contrast, OTF-NVS uses native 3DGS, whose GPU memory usage grows rapidly until offloading historical Gaussian points to disk but preventing global loop closure optimization.}
    \label{fig:GPU}
\end{minipage}
\hfill 
\begin{minipage}[t]{0.45\linewidth}
    \centering
    \includegraphics[width=\linewidth]{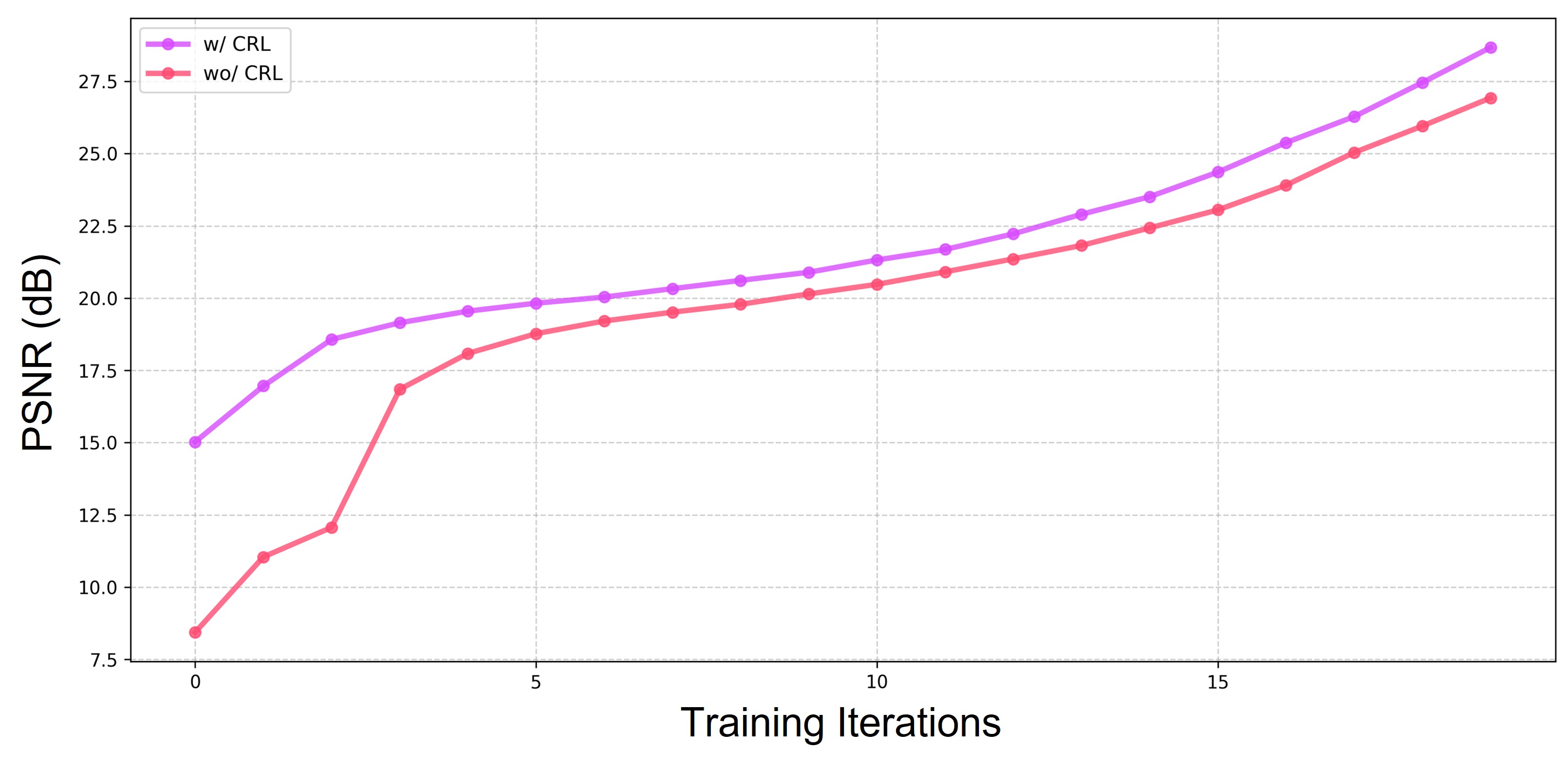}
    \caption{The rendering performance of voxelized Gaussians relative to the training iterations. CRL accelerates convergence and improves rendering quality under fixed training iterations.}
    \label{fig:ablation_color_residual_learning_2}
\end{minipage}
\end{figure*}

\paragraph{Efficiency Analysis}
Table~\ref{tab:fps_comparison} shows the runtime comparison. Our method achieves high FPS on all three datasets, maintaining real-time performance. 
The GPU memory footprint of our method comprises two components: a fixed memory footprint of less than 4 GB, which includes the MAST3R model and the shared MLP for voxelized 3DGS. The growable memory footprint increases gradually with the number of 3DGS anchors, as well as the number of keyframe nodes and edges in the factor graph. Fig.~\ref{fig:GPU} illustrates the variation in GPU memory consumption across all methods with respect to the input sequence length on the long ScanNetV2 Scene 40 sequence. Our method, which adopts Scaffold-GS, maintains low GPU memory footprint for long image sequences. In contrast, OTF-NVS uses native 3DGS, whose GPU memory usage grows rapidly until offloading historical Gaussian points to disk but preventing global loop closure optimization.

\subsection{Ablation Studies}
To validate the effectiveness of key components, we conduct ablation studies on four representative scenes shown in Table~\ref{tab:ablation}. We evaluate two variants: \textbf{wo/ SIM3} where our $\text{Sim}(3)$ global optimization is replaced with $\text{SE}(3)$ rigid transformation and \textbf{wo/ CRL} which defaults to vanilla Scaffold-GS without our color residual learning.

The results confirm that both components are critical to our method’s performance. Replacing the $\text{Sim}(3)$ optimization module leads to significant performance degradation, particularly in trajectory accuracy and tracking robustness—reduced pose precision severely impairs the rendering quality of 3DGS, verifying its essential role in mitigating cumulative pose and scale drift. Removing the CRL module results in decreased rendering fidelity and perceptual consistency, as CRL accelerates convergence within the same training iterations, whereas the variant without CRL converges more slowly, leading to inferior rendering quality.

\begin{table}
\captionsetup{skip=0pt}
\centering
\scriptsize 
\renewcommand{\arraystretch}{1} 
\caption{Performance on the Tank-and-Temples Dataset. (--: tracking failure)}
\label{tab:tank_compact}
\begin{tabular}{lcccccccc}
\hline
Method   & \multicolumn{4}{c}{Auditorium}                                     & \multicolumn{4}{c}{Ballroom}                                       \\ \cline{2-9} 
         & ATE(m)         & PSNR            & SSIM           & LPIPS          & ATE(m)         & PSNR            & SSIM           & LPIPS          \\ \hline
OTF-NVS  & -              & -               & -              & -              & 0.645          & 18.236          & 0.615          & 0.411          \\
S3PO-GS  & 1.679          & 16.019          & 0.692          & 0.726          & -              & -               & -              & -              \\
GigaSLAM & 1.026          & 22.209          & 0.930          & 0.638          & 0.201          & 20.38           & 0.898          & 0.631          \\
Ours     & \textbf{0.015} & \textbf{28.635} & \textbf{0.974} & \textbf{0.210} & \textbf{0.010} & \textbf{22.917} & \textbf{0.936} & \textbf{0.236} \\ \hline
         & \multicolumn{4}{c}{Courtroom}                                      & \multicolumn{4}{c}{Temple}                                         \\ \cline{2-9} 
         & ATE(m)         & PSNR            & SSIM           & LPIPS          & ATE(m)         & PSNR            & SSIM           & LPIPS          \\ \hline
OTF-NVS  & 0.645          & 18.66           & 0.663          & 0.455          & -              & -               & -              & -              \\
S3PO-GS  & 0.704          & 13.05           & 0.492          & 0.850          & -              & -               & -              & -              \\
GigaSLAM & 0.470          & 19.04           & 0.869          & 0.769          & -              & -               & -              & -              \\
Ours     & \textbf{0.325} & \textbf{25.192} & \textbf{0.950} & \textbf{0.230} & \textbf{0.012} & \textbf{26.254} & \textbf{0.979} & \textbf{0.174} \\ \hline
         & \multicolumn{4}{c}{Family}                                         & \multicolumn{4}{c}{Francis}                                        \\ \cline{2-9} 
         & ATE(m)         & PSNR            & SSIM           & LPIPS          & ATE(m)         & PSNR            & SSIM           & LPIPS          \\ \hline
OTF-NVS  & 0.023          & 21.22           & 0.754          & 0.300          & -              & -               & -              & -              \\
S3PO-GS  & 0.013          & 17.88           & 0.603          & 0.581          & 1.134          & 15.558          & 0.664          & 0.640          \\
GigaSLAM & 0.067          & 20.49           & 0.943          & 0.690          & -              & -               & -              & -              \\
Ours     & \textbf{0.001} & \textbf{28.130} & \textbf{0.988} & \textbf{0.106} & \textbf{0.037} & \textbf{27.397} & \textbf{0.977} & \textbf{0.199} \\ \hline
\end{tabular}
\end{table}

\begin{table}[t]
\captionsetup{skip=0pt}
\centering
\scriptsize 
\caption{Performance on the ScanNetV2 Dataset. (--: tracking failure)}
\begin{tabular}{lcccccccc}
\hline
Method   & \multicolumn{4}{c}{scene0004}                                      & \multicolumn{4}{c}{scene0005}                                      \\ \cline{2-9} 
         & ATE(m)         & PSNR            & SSIM           & LPIPS          & ATE(m)         & PSNR            & SSIM           & LPIPS          \\ \hline
OTF-NVS  & 0.543          & 24.374          & 0.824          & 0.335          & 0.415          & 23.799          & 0.866          & 0.373          \\
S3PO-GS  & -              & -               & -              & -              & 0.170          & 18.467          & 0.785          & 0.542          \\
GigaSLAM & 0.702          & 24.072          & 0.985          & 0.474          & 0.567          & 23.894          & 0.969          & 0.493          \\
Ours     & \textbf{0.098} & \textbf{30.721} & \textbf{0.997} & \textbf{0.121} & \textbf{0.085} & \textbf{30.294} & \textbf{0.990} & \textbf{0.194} \\ \hline
         & \multicolumn{4}{c}{scene0006}                                      & \multicolumn{4}{c}{scene0008}                                      \\ \cline{2-9} 
         & ATE(m)         & PSNR            & SSIM           & LPIPS          & ATE(m)         & PSNR            & SSIM           & LPIPS          \\ \hline
OTF-NVS  & 0.578          & 23.123          & 0.803          & 0.387          & 0.774          & 26.437          & 0.780          & 0.338          \\
S3PO-GS  & -              & -               & -              & -              & -              & -               & -              & -              \\
GigaSLAM & 0.458          & 22.679          & 0.960          & 0.586          & 1.132          & 26.250          & 0.980          & 0.627          \\
Ours     & \textbf{0.071} & \textbf{31.969} & \textbf{0.991} & \textbf{0.216} & \textbf{0.154} & \textbf{30.933} & \textbf{0.992} & \textbf{0.277} \\ \hline
         & \multicolumn{4}{c}{scene0010}                                      & \multicolumn{4}{c}{scene0013}                                      \\ \cline{2-9} 
         & ATE(m)         & PSNR            & SSIM           & LPIPS          & ATE(m)         & PSNR            & SSIM           & LPIPS          \\ \hline
OTF-NVS  & 0.365          & 21.662          & 0.814          & 0.347          & 0.264          & 27.385          & 0.793          & 0.392          \\
S3PO-GS  & -              & -               & -              & -              & 0.047          & 27.136          & 0.774          & 0.689          \\
GigaSLAM & 0.334          & 22.670          & 0.949          & 0.479          & 0.350          & 28.458          & \textbf{0.994} & 0.688          \\
Ours     & \textbf{0.062} & \textbf{28.615} & \textbf{0.982} & \textbf{0.252} & \textbf{0.038} & \textbf{31.755} & 0.964          & \textbf{0.404} \\ \hline
\end{tabular}
\label{tab:scannet_compact}
\end{table}

\begin{table}[t]
\captionsetup{skip=0pt}
\centering
\scriptsize 
\caption{Performance on the Waymo Dataset.}
\begin{tabular}{lcccccccc}
\hline
Method   & \multicolumn{4}{c}{seq13476}                                       & \multicolumn{4}{c}{seq132384}                                      \\ \cline{2-9} 
         & ATE(m)         & PSNR            & SSIM           & LPIPS          & ATE(m)         & PSNR            & SSIM           & LPIPS          \\ \hline
OTF-NVS  & 3.260          & 25.330          & 0.815          & 0.343          & 4.730          & 26.986          & 0.892          & 0.322          \\
S3PO-GS  & 0.612          & 21.029          & 0.719          & 0.618          & 0.774          & 18.683          & 0.825          & 0.408          \\
GigaSLAM & 2.494          & 26.148          & 0.972          & 0.530          & 0.958          & 26.267          & 0.987          & 0.475          \\
Ours     & \textbf{0.426} & \textbf{29.436} & \textbf{0.986} & \textbf{0.185} & \textbf{0.463} & \textbf{29.570} & \textbf{0.977} & \textbf{0.299} \\ \hline
         & \multicolumn{4}{c}{seq152706}                                      & \multicolumn{4}{c}{seq158686}                                      \\ \cline{2-9} 
         & ATE(m)         & PSNR            & SSIM           & LPIPS          & ATE(m)         & PSNR            & SSIM           & LPIPS          \\ \hline
OTF-NVS  & 1.975          & 27.454          & 0.855          & 0.345          & 1.682          & \textbf{28.193} & 0.872          & \textbf{0.267} \\
S3PO-GS  & 1.136          & 21.892          & 0.790          & 0.574          & 1.296          & 20.096          & 0.729          & 0.520          \\
GigaSLAM & 0.528          & 27.072          & 0.973          & 0.411          & 0.614          & 25.856          & 0.973          & 0.460          \\
Ours     & \textbf{0.517} & \textbf{31.593} & \textbf{0.987} & \textbf{0.216} & \textbf{0.446} & 23.269          & \textbf{0.948} & 0.429          \\ \hline
         & \multicolumn{4}{c}{seq163453}                                      & \multicolumn{4}{c}{seq405841}                                      \\ \cline{2-9} 
         & ATE(m)         & PSNR            & SSIM           & LPIPS          & ATE(m)         & PSNR            & SSIM           & LPIPS          \\ \hline
OTF-NVS  & 6.141          & 24.639          & 0.825          & 0.352          & 14.707         & 25.769          & 0.834          & 0.374          \\
S3PO-GS  & \textbf{0.935} & 21.185          & 0.751          & 0.561          & 0.583          & 22.801          & 0.800          & 0.479          \\
GigaSLAM & 5.213          & 24.080          & 0.978          & 0.490          & 0.750          & 28.290          & 0.973          & 0.411          \\
Ours     & 0.942          & \textbf{27.182} & \textbf{0.977} & \textbf{0.299} & \textbf{0.577} & \textbf{29.145} & \textbf{0.976} & \textbf{0.206} \\ \hline
\end{tabular}
\label{tab:waymo_compact}
\end{table}

\begin{table}[h]
\captionsetup{skip=0pt}
\scriptsize 
\caption{FPS comparison across different datasets. Noted that FPS is correlated with the density of selected video keyframes.}
\begin{tabular}{lllll}
\hline
Dataset/Scene & OTF-NVS & S3PO-GS & GigaSLAM & Ours  \\ \hline
ScanNetV2              & 6.439            & 0.297            & 1.073             & \textbf{7.067} \\ 
TanksTemples           & \textbf{1.706}   & 0.197            & 0.426             & 1.279          \\ 
Waymo                  & 3.316            & 0.094            & 0.829             & \textbf{4.320} \\ \hline
\end{tabular}
\label{tab:fps_comparison}
\end{table}

\begin{table}[t]
\captionsetup{skip=0pt}
\centering
\scriptsize 
\renewcommand{\arraystretch}{0.8} 
\caption{Ablation Studies.}
\begin{tabular}{l *{8}{c}}
\toprule
Method & \multicolumn{4}{c}{scene0010} & \multicolumn{4}{c}{scene0040} \\
\cmidrule(lr){2-5} \cmidrule(lr){6-9}
& ATE(m) & PSNR & SSIM & LPIPS & ATE(m) & PSNR & SSIM & LPIPS \\
\midrule
wo/ SIM3 & 0.062 & 26.346 & 0.972 & 0.328 & 0.111 & 26.033 & 0.981 & 0.305 \\
wo/ CRL & \textbf{0.058} & 26.957 & 0.975 & 0.321 & \textbf{0.069} & 26.723 & 0.984 & 0.274 \\
Ours & \textbf{0.058} & \textbf{28.615} & \textbf{0.982} & \textbf{0.252} & \textbf{0.069} & \textbf{28.660} & \textbf{0.990} & \textbf{0.159} \\
\midrule
Method & \multicolumn{4}{c}{Playground} & \multicolumn{4}{c}{Auditorium} \\
\cmidrule(lr){2-5} \cmidrule(lr){6-9}
& ATE(m) & PSNR & SSIM & LPIPS & ATE(m) & PSNR & SSIM & LPIPS \\
\midrule
wo/ SIM3 & 0.115 & 22.417 & 0.960 & 0.454 & 0.245 & 24.953 & 0.951 & 0.450 \\
wo/ CRL & \textbf{0.019} & 24.107 & 0.972 & 0.335 & \textbf{0.015} & 26.982 & 0.965 & 0.249 \\
Ours & \textbf{0.019} & \textbf{24.952} & \textbf{0.975} & \textbf{0.288} & \textbf{0.015} & \textbf{28.635} & \textbf{0.974} & \textbf{0.210} \\
\bottomrule
\end{tabular}%
\label{tab:ablation}
\end{table}

\section{Applications}
\label{sec:Applications}

\subsection{UAV Active Reconstruction}
We apply our method to unmanned aerial vehicles (UAVs) active reconstruction~\cite{chen2011active}, where the UAV is required not only to perform 3D reconstruction using captured images but also to autonomously plan and execute collision-free moving trajectories to capture new images simultaneously. Our method can obtain high-quality voxelized Gaussian maps in real time, enabling monocular UAVs to plan paths based on voxel grids during reconstruction. As shown in the Fig.~\ref{fig:teaser}, our UAV is equipped with a single camera. The UAV captures image streams via the camera and transmits them to the server-side our method through WIFI, which runs on an L40 GPU server.
On the server side, we first reconstruct the voxelized Gaussian map from the images. Subsequently, to plan the path for the UAV, we first mark the voxels within the rendered depth range as navigable areas to obtain an occupancy map with a resolution set to $0.2m^3$. We then sample some viewpoints within the navigable areas and render 3DGS. The next best view (NBV) is selected by comprehensively considering two criteria: poor 3DGS rendering completeness judged by opacity renderings and low travel distance. We further employ the A*~\cite{hart1968formal} algorithm on the occupancy map to generate a collision-free path, which is then sent back to the UAV via WIFI for path execution and new image capture. Safe flight is achieved without the need for explicitly deploying additional obstacle avoidance algorithms in static scenes.
As our method with only image input lacks access to real-world scale, we use a visual odometer on the UAV to obtain real-world scale. During the initialization phase, we align our estimated poses of the first 16 keyframes with the odometer poses with the real scale using the Umeyama algorithm~\cite{umeyama2002eigendecomposition}. This ensures that the Gaussian reconstruction results can represent 3D scene with real scale. For detailed and comprehensive illustration of the active reconstruction process, please refer to Part 3 of the supplementary material video.\looseness=-1

\subsection{Other Potential Applications}

Our online real-time Gaussian reconstruction are promising to be applied to more robotics applications, such as planning, navigation, AR/VR, and scene semantic understanding. 
Additionally, it can support user-friendly mobile applications on any camera devices, where users can perform online Gaussian reconstruction while freely observing new viewpoints and monitoring the reconstruction progress. This assists users in completing reconstruction tasks efficiently, as opposed to the traditional pipeline (e.g., COLMAP~\cite{schonberger2016structure}) combined with 3DGS, which requires hours of offline processing and suffers from a relatively high failure rate for casual shot videos captured by non-professional users.

\section{Limitations}
\label{sec:limitations}

Our method does not consider dynamic objects. Minor dynamic objects in the scene are overlooked; however, when the proportion of dynamic objects exceeds a certain threshold, it degrades pose estimation and multi-view consistent rendering of the static 3D Gaussian Splatting model. Specifically, dynamic objects introduce mismatches in the predicted pointmaps, which further lead to incorrect camera pose estimation and artifacts in the Gaussian map.
On the other hand, since our two-view geometric estimation relies on MAST3R, the confidence of the pointmap predicted by MAST3R decreases with increasing distance. This limitation results in a significant drop in accuracy for some outdoor scenes.
Furthermore, while our current method can handle thousands of images, the GPU memory required by our method still increases with the length of the input video as the number of edges in the factor graph grows proportionally with the number of nodes. Due to the necessity for simultaneous $\text{Sim}(3)$ global optimization, our method cannot adopt the anchor-based strategy used in OTF-NVS to offload historical 3DGS data to the disk, restricting its scalability for extremely long video sequences. \looseness=-1

\section{Conclusion}
\label{sec:conclusion}

This paper proposes an efficient and robust online 3DGS method. Our method addresses two key limitations of prior online 3DGS approaches: fragile camera tracking due to the lack of global optimization and inefficient optimization in large-scale scenes. By integrating a $\text{Sim}(3)$ global optimization module, we jointly refine camera poses and the voxelized 3D Gaussian map, enabling reliable loop closure and consistent reconstruction over long sequences. Furthermore, our proposed color residual learning strategy accelerates the convergence of the voxelized 3DGS representation while improving rendering quality.
Extensive experiments on diverse indoor and outdoor scenes validate the superiority of our method. We also develop and deploy a practical UAV-based active reconstruction system, showcasing the robustness and applicability of our method in real-world scenarios.

\begin{acks}
This work was supported by National Natural Science Foundation of China (Grant No.: 62372015), Leading Projects in Key Research Fields of Language Funded by the National Language Commission, Key Laboratory of Intelligent Press Media Technology, and National Engineering Research Center of New Electronic Publishing Technologies.
\end{acks}

\bibliographystyle{ACM-Reference-Format}
\bibliography{ref}

@inproceedings{sun2020scalability,
  title={Scalability in perception for autonomous driving: Waymo open dataset},
  author={Sun, Pei and Kretzschmar, Henrik and Dotiwalla, Xerxes and Chouard, Aurelien and Patnaik, Vijaysai and Tsui, Paul and Guo, James and Zhou, Yin and Chai, Yuning and Caine, Benjamin and others},
  booktitle={Proceedings of the IEEE/CVF conference on computer vision and pattern recognition},
  pages={2446--2454},
  year={2020}
}

@inproceedings{wen2025segs,
  title={Segs-slam: Structure-enhanced 3d gaussian splatting slam with appearance embedding},
  author={Wen, Tianci and Liu, Zhiang and Fang, Yongchun},
  booktitle={Proceedings of the IEEE/CVF International Conference on Computer Vision},
  pages={28103--28113},
  year={2025}
}

@article{zhang2025hi,
  title={Hi-slam2: Geometry-aware gaussian slam for fast monocular scene reconstruction},
  author={Zhang, Wei and Cheng, Qing and Skuddis, David and Zeller, Niclas and Cremers, Daniel and Haala, Norbert},
  journal={IEEE Transactions on Robotics},
  volume={41},
  pages={6478--6493},
  year={2025},
  publisher={IEEE}
}

@article{levenberg1944method,
  title={A method for the solution of certain non-linear problems in least squares},
  author={Levenberg, Kenneth},
  journal={Quarterly of applied mathematics},
  volume={2},
  number={2},
  pages={164--168},
  year={1944}
}

@inproceedings{tolias2013aggregate,
  title={To aggregate or not to aggregate: Selective match kernels for image search},
  author={Tolias, Giorgos and Avrithis, Yannis and J{\'e}gou, Herv{\'e}},
  booktitle={Proceedings of the IEEE international conference on computer vision},
  pages={1401--1408},
  year={2013}
}

@inproceedings{duisterhof2025mast3r,
  title={Mast3r-sfm: a fully-integrated solution for unconstrained structure-from-motion},
  author={Duisterhof, Bardienus Pieter and Zust, Lojze and Weinzaepfel, Philippe and Leroy, Vincent and Cabon, Yohann and Revaud, Jerome},
  booktitle={2025 International Conference on 3D Vision (3DV)},
  pages={1--10},
  year={2025},
  organization={IEEE}
}

@article{hart1968formal,
  title={A formal basis for the heuristic determination of minimum cost paths},
  author={Hart, Peter E and Nilsson, Nils J and Raphael, Bertram},
  journal={IEEE transactions on Systems Science and Cybernetics},
  volume={4},
  number={2},
  pages={100--107},
  year={1968},
  publisher={IEEE}
}

@inproceedings{zhu2025loopsplat,
  title={Loopsplat: Loop closure by registering 3d gaussian splats},
  author={Zhu, Liyuan and Li, Yue and Sandstr{\"o}m, Erik and Huang, Shengyu and Schindler, Konrad and Armeni, Iro},
  booktitle={2025 International Conference on 3D Vision (3DV)},
  pages={156--167},
  year={2025},
  organization={IEEE}
}

@article{lin2025depth,
  title={Depth anything 3: Recovering the visual space from any views},
  author={Lin, Haotong and Chen, Sili and Liew, Junhao and Chen, Donny Y and Li, Zhenyu and Shi, Guang and Feng, Jiashi and Kang, Bingyi},
  journal={arXiv preprint arXiv:2511.10647},
  year={2025}
}

@article{deng2025vggt,
  title={VGGT-Long: Chunk it, Loop it, Align it--Pushing VGGT's Limits on Kilometer-scale Long RGB Sequences},
  author={Deng, Kai and Ti, Zexin and Xu, Jiawei and Yang, Jian and Xie, Jin},
  journal={arXiv preprint arXiv:2507.16443},
  year={2025}
}

@article{davison2007monoslam,
  title={MonoSLAM: Real-time single camera SLAM},
  author={Davison, Andrew J and Reid, Ian D and Molton, Nicholas D and Stasse, Olivier},
  journal={IEEE transactions on pattern analysis and machine intelligence},
  volume={29},
  number={6},
  pages={1052--1067},
  year={2007},
  publisher={IEEE}
}

@article{umeyama2002eigendecomposition,
  title={An eigendecomposition approach to weighted graph matching problems},
  author={Umeyama, Shinji},
  journal={IEEE transactions on pattern analysis and machine intelligence},
  volume={10},
  number={5},
  pages={695--703},
  year={2002},
  publisher={IEEE}
}

@inproceedings{schonberger2016structure,
  title={Structure-from-motion revisited},
  author={Schonberger, Johannes L and Frahm, Jan-Michael},
  booktitle={Proceedings of the IEEE conference on computer vision and pattern recognition},
  pages={4104--4113},
  year={2016}
}

@article{chen2011active,
  title={Active vision in robotic systems: A survey of recent developments},
  author={Chen, Shengyong and Li, Youfu and Kwok, Ngai Ming},
  journal={The International Journal of Robotics Research},
  volume={30},
  number={11},
  pages={1343--1377},
  year={2011},
  publisher={Sage Publications Sage UK: London, England}
}

@inproceedings{tolias2020learning,
  title={Learning and aggregating deep local descriptors for instance-level recognition},
  author={Tolias, Giorgos and Jenicek, Tomas and Chum, Ond{\v{r}}ej},
  booktitle={European Conference on Computer Vision},
  pages={460--477},
  year={2020},
  organization={Springer}
}

@inproceedings{deng2025gigaslam,
  title={Gigaslam: Large-scale monocular slam with hierarchical gaussian splats},
  author={Deng, Kai and Zhang, Yigong and Yang, Jian and Xie, Jin},
  booktitle={Proceedings of the SIGGRAPH Asia 2025 Conference Papers},
  pages={1--10},
  year={2025}
}

@inproceedings{cheng2025outdoor,
  title={Outdoor monocular slam with global scale-consistent 3d gaussian pointmaps},
  author={Cheng, Chong and Yu, Sicheng and Wang, Zijian and Zhou, Yifan and Wang, Hao},
  booktitle={Proceedings of the IEEE/CVF International Conference on Computer Vision},
  pages={26035--26044},
  year={2025}
}

@article{knapitsch2017tanks,
  title={Tanks and temples: Benchmarking large-scale scene reconstruction},
  author={Knapitsch, Arno and Park, Jaesik and Zhou, Qian-Yi and Koltun, Vladlen},
  journal={ACM Transactions on Graphics (ToG)},
  volume={36},
  number={4},
  pages={1--13},
  year={2017},
  publisher={ACM New York, NY, USA}
}

@inproceedings{wang2025vggt,
  title={Vggt: Visual geometry grounded transformer},
  author={Wang, Jianyuan and Chen, Minghao and Karaev, Nikita and Vedaldi, Andrea and Rupprecht, Christian and Novotny, David},
  booktitle={Proceedings of the Computer Vision and Pattern Recognition Conference},
  pages={5294--5306},
  year={2025}
}

@inproceedings{murai2025mast3r,
  title={MASt3R-SLAM: Real-time dense SLAM with 3D reconstruction priors},
  author={Murai, Riku and Dexheimer, Eric and Davison, Andrew J},
  booktitle={Proceedings of the Computer Vision and Pattern Recognition Conference},
  pages={16695--16705},
  year={2025}
}

@article{maggio2025vggt,
  title={Vggt-slam: Dense rgb slam optimized on the sl (4) manifold},
  author={Maggio, Dominic and Lim, Hyungtae and Carlone, Luca},
  journal={arXiv preprint arXiv:2505.12549},
  year={2025}
}

@article{li2025artdeco,
  title={Artdeco: Towards efficient and high-fidelity on-the-fly 3d reconstruction with structured scene representation},
  author={Li, Guanghao and Ren, Kerui and Xu, Linning and Zheng, Zhewen and Jiang, Changjian and Gao, Xin and Dai, Bo and Pu, Jian and Yu, Mulin and Pang, Jiangmiao},
  journal={arXiv preprint arXiv:2510.08551},
  year={2025}
}

@inproceedings{barron2021mipnerf,
  author    = {Jonathan T Barron and Ben Mildenhall and Matthew Tancik and Peter Hedman and Ricardo Martin-Brualla and Pratul P Srinivasan},
  title     = {Mip-nerf: A multiscale representation for anti-aliasing neural radiance fields},
  booktitle = {Proceedings of the IEEE/CVF International Conference on Computer Vision},
  pages     = {5855--5864},
  year      = {2021}
}

@inproceedings{dai2017scannet,
  author    = {Angela Dai and Angel X. Chang and Manolis Savva and Maciej Halber and Thomas Funkhouser and Matthias Nie{\ss}ner},
  title     = {Scannet: Richly-annotated 3d reconstructions of indoor scenes},
  booktitle = {Proc. Computer Vision and Pattern Recognition (CVPR)},
  year      = {2017},
  publisher = {IEEE}
}

@inproceedings{keetha2024splatam,
  author    = {Nikhil Keetha and Jay Karhade and Krishna Murthy Jatavallabhula and Gengshan Yang and Sebastian Scherer and Deva Ramanan and Jonathon Luiten},
  title     = {Splatam: Splat, track \& map 3d gaussians for dense rgb-d slam},
  booktitle = {Proceedings of the IEEE/CVF Conference on Computer Vision and Pattern Recognition},
  year      = {2024}
}

@article{kerbl2023,
  author    = {Bernhard Kerbl and Georgios Kopanas and Thomas Leimk{\"u}hler and George Drettakis},
  title     = {3d gaussian splatting for real-time radiance field rendering},
  journal   = {ACM Transactions on Graphics},
  volume    = {42},
  number    = {4},
  pages     = {1--14},
  year      = {2023}
}

@misc{leroy2024,
  author    = {Vincent Leroy and Yohann Cabon and Jerome Revaud},
  title     = {Grounding Image Matching in 3D with MASt3R},
  year      = {2024}
}

@article{li2026ecslam,
  author    = {Guanghao Li and Qi Chen and Yuxiang Yan and Jian Pu},
  title     = {Ec-slam: Effectively constrained neural rgb-d slam with tsdf hash encoding and joint optimization},
  journal   = {Pattern Recognition},
  volume    = {170},
  pages     = {112034},
  year      = {2026}
}

@inproceedings{matsuki2024gaussian,
  author    = {Hidenobu Matsuki and Riku Murai and Paul HJ Kelly and Andrew J Davison},
  title     = {Gaussian splatting slam},
  booktitle = {Proceedings of the IEEE/CVF Conference on Computer Vision and Pattern Recognition},
  year      = {2024}
}

@article{meuleman2025,
  author    = {Andreas Meuleman and Ishaan Shah and Alexandre Lanvin and Bernhard Kerbl and George Drettakis},
  title     = {On-the-fly reconstruction for large-scale novel view synthesis from unposed images},
  journal   = {ACM Transactions on Graphics (TOG)},
  volume    = {44},
  number    = {4},
  pages     = {1--14},
  year      = {2025}
}

@misc{murai2024mast3rslam,
  author    = {Riku Murai and Eric Dexheimer and Andrew J. Davison},
  title     = {MASt3R-SLAM: Real-time dense SLAM with 3D reconstruction priors},
  year      = {2024},
  eprint    = {arXiv preprint}
}

@inproceedings{sucar2021imap,
  author    = {Edgar Sucar and Shikun Liu and Joseph Ortiz and Andrew J Davison},
  title     = {imap: Implicit mapping and positioning in real-time},
  booktitle = {Proceedings of the IEEE/CVF international conference on computer vision},
  pages     = {6229--6238},
  year      = {2021}
}

@inproceedings{teed2021droid,
  author    = {Zachary Teed and Jia Deng},
  title     = {Droid-slam: Deep visual slam for monocular, stereo, and rgb-d cameras},
  booktitle = {Advances in neural information processing systems},
  volume    = {34},
  pages     = {16558--16569},
  year      = {2021}
}

@book{hartley2003multiple,
  title={Multiple view geometry in computer vision},
  author={Hartley, Richard and Zisserman, Andrew},
  year={2003},
  publisher={Cambridge university press}
}

@inproceedings{zhang2023go,
  author    = {Youmin Zhang and Fabio Tosi and Stefano Mattoccia and Matteo Poggi},
  title     = {Go-slam: Global optimization for consistent 3d instant reconstruction},
  booktitle = {Proceedings of the IEEE/CVF International Conference on Computer Vision},
  pages     = {3727--3737},
  year      = {2023}
}

@inproceedings{zhu2022nice,
  author    = {Zihan Zhu and Songyou Peng and Viktor Larsson and Weiwei Xu and Hujun Bao and Zhaopeng Cui and Martin R. Oswald and Marc Pollefeys},
  title     = {Nice-slam: Neural implicit scalable encoding for slam},
  booktitle = {Proceedings of the IEEE/CVF Conference on Computer Vision and Pattern Recognition (CVPR)},
  year      = {2022}
}

@inproceedings{bloesch2018codeslam,
  author    = {Michael Bloesch and Jan Czarnowski and Ronald Clark and Stefan Leutenegger and Andrew J Davison},
  title     = {Codeslam-learning a compact, optimisable representation for dense visual slam},
  booktitle = {Proceedings of the IEEE conference on computer vision and pattern recognition},
  pages     = {2560--2568},
  year      = {2018}
}

@article{cadena2016past,
  author    = {Cesar Cadena and Luca Carlone and Henry Carrillo and Yasir Latif and Davide Scaramuzza and Jos{\'e} Neira and Ian Reid and John J Leonard},
  title     = {Past, present, and future of simultaneous localization and mapping: Toward the robust-perception age},
  journal   = {IEEE Transactions on robotics},
  volume    = {32},
  number    = {6},
  pages     = {1309--1332},
  year      = {2016}
}

@article{campos2021orb3,
  author    = {Carlos Campos and Richard Elvira and Juan J. Gomez Rodriguez and Jose M. M. Montiel and Juan D. Tardos},
  title     = {ORB-SLAM3: An accurate open-source library for visual, visual--inertial, and multimap SLAM},
  journal   = {IEEE Transactions on Robotics},
  volume    = {37},
  number    = {6},
  pages     = {1874--1890},
  year      = {2021}
}

@inproceedings{lindenberger2021pixel,
  author    = {Philipp Lindenberger and Paul-Edouard Sarlin and Viktor Larsson and Marc Pollefeys},
  title     = {Pixel-perfect structure-from-motion with featuremetric refinement},
  booktitle = {Proceedings of the IEEE/CVF International Conference on Computer Vision},
  pages     = {5987--5997},
  year      = {2021}
}

@inproceedings{lu2024scaffold,
  author    = {Tao Lu and Mulin Yu and Linning Xu and Yuanbo Xiangli and Limin Wang and Dahua Lin and Bo Dai},
  title     = {Scaffold-gs: Structured 3d gaussians for view-adaptive rendering},
  booktitle = {Proceedings of the IEEE/CVF Conference on Computer Vision and Pattern Recognition},
  pages     = {20654--20664},
  year      = {2024}
}

@article{murartal2017orb2,
  author    = {Raul Mur-Artal and Juan D. Tardos},
  title     = {ORB-SLAM2: An open-source SLAM system for monocular, stereo, and RGBd cameras},
  journal   = {IEEE Transactions on Robotics},
  volume    = {33},
  number    = {5},
  pages     = {1255--1262},
  year      = {2017}
}

@article{murartal2015orb,
  author    = {Raul Mur-Artal and J. M. M. Montiel and Juan D. Tardos},
  title     = {ORBSLAM: A versatile and accurate monocular SLAM system},
  journal   = {IEEE Transactions on Robotics},
  volume    = {31},
  number    = {5},
  pages     = {1147--1163},
  year      = {2015}
}

@misc{sandstrom2024splat,
  author    = {Erik Sandstr{\"o}m and Keisuke Tateno and Michael Oechsle and Michael Niemeyer and Luc Van Gool and Martin R Oswald and Federico Tombari},
  title     = {Splat-slam: Globally optimized rgb-only slam with 3d gaussians},
  year      = {2024},
  eprint    = {2405.16544},
  archivePrefix = {arXiv},
  primaryClass = {cs.CV}
}

@misc{tang2018ba,
  author    = {Chengzhou Tang and Ping Tan},
  title     = {Ba-net: Dense bundle adjustment network},
  year      = {2018},
  eprint    = {1806.04807},
  archivePrefix = {arXiv},
  primaryClass = {cs.CV}
}

@inproceedings{wang2024dust3r,
  author    = {Shuzhe Wang and Vincent Leroy and Yohann Cabon and Boris Chidlovskii and Jerome Revaud},
  title     = {Dust3r: Geometric 3d vision made easy},
  booktitle = {Proceedings of the IEEE/CVF Conference on Computer Vision and Pattern Recognition},
  pages     = {20697--20709},
  year      = {2024}
}

@inproceedings{yan2024gs,
  author    = {Chi Yan and Delin Qu and Dan Xu and Bin Zhao and Zhigang Wang and Dong Wang and Xuelong Li},
  title     = {Gs-slam: Dense visual slam with 3d gaussian splatting},
  booktitle = {Proceedings of the IEEE/CVF Conference on Computer Vision and Pattern Recognition},
  pages     = {19595--19604},
  year      = {2024}
}

\clearpage

\begin{figure*}[!t]
\captionsetup{skip=0pt}
\centering
\includegraphics[width=0.7\linewidth]{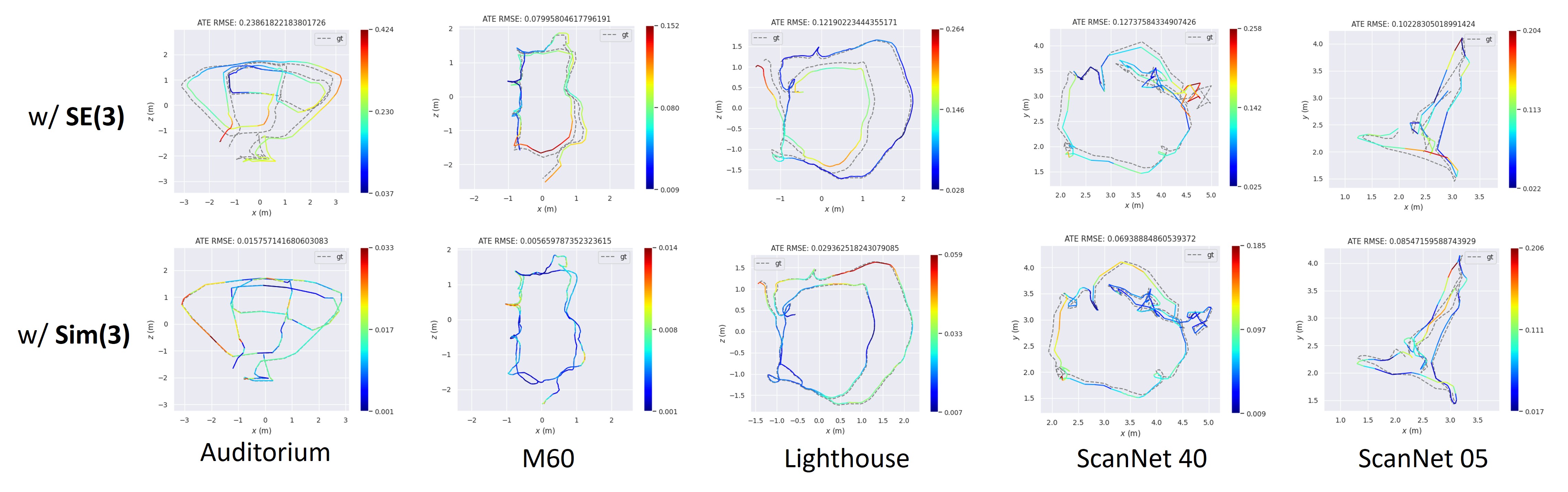}
\caption{\textbf{$\text{SE}(3)$} denotes the use of rigid transformations only, which fails to address the cumulative scale errors of 3DGS anchors caused by the inherent scale ambiguity of visual depth prediction models. In contrast, \textbf{$\text{Sim}(3)$} refers to global loop closure adjustment incorporating both rigid transformations and scale optimizations. This enables more effective global scale consistency optimization of 3DGS anchors, thereby yielding a substantial performance improvement—particularly in trajectory accuracy, global 3DGS loop consistency, and rendering quality.}
\label{fig:ablation_SIM3}
\end{figure*}

\end{document}